\documentclass[a4paper, 10pt, twocolumn, twoside]{article}
\usepackage{amssymb}

\usepackage{ISARC}

\usepackage{lscape}
\usepackage{hologo}
\usepackage[none]{hyphenat}
\def\mathbi#1{\textbf{\em #1}}

\begin{document}

 % Do not change the following line
\linespread{0.5}
\setlength{\tabcolsep}{4pt} % Default value: 6pt

\title{CoNav Chair: Design of a ROS-based Smart Wheelchair for Shared Control Navigation in the Built Environment}

\author{Yifan Xu$^1$, Qianwei Wang$^2$, Jordan Lillie$^3$, Vineet Kamat$^4$ and Carol Menassa$^5$}

\affiliation{
$^1$Department of Civil and Environmental Engineering, University of Michigan, United States of America\\
$^2$College of Literature, Science, and the Arts, University of Michigan, United States of America\\
$^3$Wheelchair Seating Service, University of Michigan Health, United States of America\\
$^4$Department of Civil and Environmental Engineering, University of Michigan, United States of America\\
$^5$Department of Civil and Environmental Engineering, University of Michigan, United States of America
}

\email{
\href{mailto:e.yfx@umich.edu}{yfx@umich.edu}, 
\href{mailto:e.qweiw@umich.edu}{qweiw@umich.edu},
\href{mailto:e.jlillie@med.umich.edu}{jlillie@med.umich.edu},
\href{mailto:e.vkamat@umich.edu}{vkamat@umich.edu},
\href{mailto:e.menassa@umich.edu}{menassa@umich.edu}
}

% Do not change the following three lines
\maketitle 
\thispagestyle{fancy} 
\pagestyle{fancy}

\begin{abstract}
With the number of people with disabilities (PWD) increasing worldwide each year, the demand for mobility support to enable independent living and social integration is also growing. Wheelchairs commonly support the mobility of PWD in both indoor and outdoor environments. However, current powered wheelchairs (PWC) often fail to meet the needs of PWD, who may find it difficult to operate them. Moreover, existing research on robotic wheelchairs typically focuses either on full autonomy or enhanced manual control, which can lead to reduced efficiency and user trust. To address these issues, this paper proposes a Robot Operating System (ROS)-based smart wheelchair (CoNav Chair) that incorporates a shared control navigation algorithm and obstacle avoidance to support PWD while fostering efficiency and trust between the robot and the user. Our design is divided into hardware and software components. Experimental results conducted in a typical indoor social environment demonstrate the performance and effectiveness of our smart wheelchair’s hardware and software design. This integrated design fosters trust and autonomy, crucial for the acceptance of assistive mobility technologies in the built environment.

\end{abstract}

\begin{keywords}
Smart Wheelchair; Navigation; Shared Control 
\end{keywords}

\section{Introduction}
\label{sec:Introduction}
\begin{figure*}[h!]
    \centering
    \includegraphics[width = \textwidth]{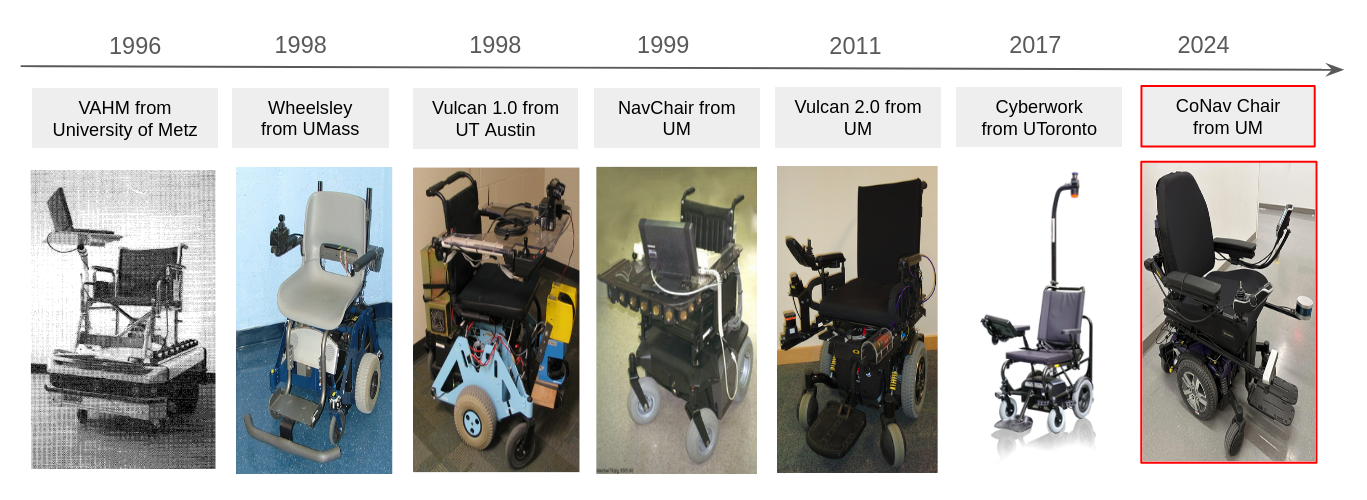}
    \caption{Development Timeline of Smart Wheelchairs}
    \label{fig:previous work}
\end{figure*}
% \subsection{Background and Challenge}
With the number of people with disabilities (PWD) increasing annually, more and more people will have to rely on wheelchairs to facilitate their mobility to live their lives independently. In the United States, nearly 2.7 million people experience mobility challenges leading them to rely on wheelchairs in their daily lives~\cite{Koontz2015Wheeled}. That is expected to increase by 7\% annually~\cite{Flagg2009Chapter1} leading to a growing demand for mobility support systems that enable independent living and social integration. Wheelchairs, especially powered wheelchairs (PWC), are the most common means of assisting PWD in both indoor and outdoor environments where these devices provide crucial mobility assistance, allowing users to navigate their surroundings more easily~\cite{Cooper1998Wheelchair,Minkel2002Seating}. In addition, smart wheelchairs created to introduce autonomy in PWC operation have demonstrated significant potential in improving the quality of life for PWD~\cite{Fehr2000Adequacy}. 

Current smart wheelchairs face a notable limitation: they often focus exclusively on either full autonomy or enhanced manual control~\cite{Fehr2000Adequacy}. This imbalance can result in inefficiencies during navigation and a lack of trust in the technology—critical factors that influence the successful adoption of these devices~\cite{Argyros2004}. Consequently, these challenges hinder user acceptance and integration into daily life~\cite{Sakakibara2013Wheelchair}. Furthermore, the limited field of view of sensors and the suboptimal performance of autonomous navigation algorithms in avoiding obstacles can leave users feeling unsafe and wishing to take control. To address these concerns, a shared control-based smart wheelchair offers a potential solution. Shared control enables users to partially override the autonomous navigation system, allowing them to contribute to decision-making during navigation. This approach not only enhances the mobility of PWD but also fosters user trust by offering a balanced partnership between autonomy and manual control.

% Building on this concept, we developed Co-Nav Chair, a shared control-based smart wheelchair that combines advanced hardware and software design to address these challenges.

% \subsection{Research Objectives and Contributions}
Building on insights from previous smart wheelchair designs~\cite{Bourhis1996Mobile,Gribble1998Integrating,Yanco1998Wheelesley,Levine1999NavChair,Park2017DiscreteTime,Kopun2023Toronto} and shared control-based navigation structures~\cite{xu2024}, we developed both the hardware and software for such a shared control-based smart wheelchair that we call the CoNav Chair.

For hardware design, We based our hardware design on a widely used commercial PWC. The key elements include the integration of LiDAR, a camera, an inertial measurement unit (IMU), and two encoders. These sensors collectively provide comprehensive input data for the system. To accommodate all the sensors and the controller board, we designed a dedicated sensor mounting board located under the wheelchair. This sophisticated hardware integration lays the foundation for the implementation of the advanced shared control-based navigation algorithm, ultimately contributing to a more intuitive, reliable, and efficient mobility solution for people with disabilities. Additionally, we implemented a closed-loop proportional–integral–derivative (PID)~\cite{KiamHeongAng2005} motor control module to manage the wheelchair's motors effectively. 

The software design centers around a ROS-based Simultaneous Localization and Mapping (SLAM) and navigation framework, which facilitates navigation and interactive motor control. For the SLAM module, we employ Faster-LIO~\cite{Bai2022FasterLIO} to estimate the wheelchair's states and positions and RTAB-Map~\cite{Labbe2013AppearanceBased} to construct a detailed map of the indoor environment. This 2D map is then used in our shared control-based navigation system. By integrating the user's control input signals with the autonomous navigation system, the wheelchair can navigate effectively through the mapped area. 
% On the hardware side, we created a differential drive-based wheelchair model equipped with multiple integrated sensors, including LiDAR, a camera, an IMU, and encoders. This multi-sensor setup enables the wheelchair to perceive its surroundings in real-time, facilitating precise and responsive movements. On the software side, we implemented a shared control-based navigation system~\cite{xu2024} that allows the wheelchair to navigate effectively while adapting to the user's control inputs. This system enhances the interaction between the user and the robot, fostering greater trust and cooperation. Generally speaking, this hardware and software design solved two parts of the challenge from the side of PWD and smart wheelchair development.

In general, our proposed system has the following contributions:
\begin{itemize}
    \item We developed the CoNav Chair, a shared control-based smart wheelchair platform that bridges the gap between full autonomy and manual control by leveraging our proposed shared control algorithm, setting a new benchmark for assistive mobility solutions.
    \item At the hardware level, we developed a ROS-based smart wheelchair hardware platform built from a widely used commercial PWC with essential control electronics and sensors to enable advanced perception, mapping, autonomous navigation, and precise motor control.
    \item At the software level, we proposed a novel shared control-based navigation framework for assistive mobile robots integrating a user’s desire for path preference and control during navigation to improve the efficiency and human-robot trust.
    \item We have tested our wheelchair system in a built environment and demonstrated that our robot system outperforms other autonomous wheelchair navigation systems both in effectiveness and user comfort. 
\end{itemize}

\section{Related Work}
\label{sec:relatedwork}
\par 

\begin{table*}
\centering
\caption{Comparison between the previous wheelchair models and the CoNav Chair}
\renewcommand{\arraystretch}{2} % Adjust row spacing (default: 1.0)
\footnotesize
\label{tab: wheelchair comparison}
\begin{tabular}{ccccc}
\hline
\textbf{Wheelchair Models}&\textbf{SLAM compatible} & \textbf{Autonomous navigation} & \textbf{Shared control by user} & \textbf{Sensors}\\
\hline
Powered wheelchair	& N/A & N/A & N/A & N/A \\
VAHM~\cite{Bourhis1996Mobile}	& N/A & N/A & N/A & Contact, Ultrasonic sensor \\
Wheelsley~\cite{Yanco1998Wheelesley}	& N/A & N/A & \cmark & Eye tracking system \\
Vulcan 1.0~\cite{Gribble1998Integrating}	& \cmark & N/A & N/A & 2D LiDAR and camera \\
Nav-Chair~\cite{Levine1999NavChair} & N/A &\cmark & N/A & 12 ultrasonic sensors \\
Vulcan 2.0~\cite{Park2017DiscreteTime}	& \cmark & \cmark & N/A & 2D LiDAR \\
Cyberwork~\cite{Kopun2023Toronto} & \cmark & \cmark & N/A & 3 Cameras and eye \\
\textbf{Co-Nav Chair}(Ours) & \cmark & \cmark & \cmark & 3D LiDAR and RGB-D camera \\
\hline
\end{tabular}
\end{table*}

The development of smart wheelchairs has been a focal point in the literature aimed at improving the mobility and independence of PWD. A lot of research on wheelchair hardware instrumentation and software integration has been conducted and implemented to address the limitations of the traditional PWC.

The research on smart wheelchairs started in the 1990s, and a lot of work related to smart wheelchairs has been developed such as VAHM~\cite{Bourhis1996Mobile}, Vulcan1.0~\cite{Gribble1998Integrating}, Wheelsley~\cite{Yanco1998Wheelesley} and NavChair~\cite{Levine1999NavChair}, Vulcan2.0~\cite{Park2017DiscreteTime} and Cyberwork Wheelchair~\cite{Kopun2023Toronto}. The development timeline of the smart wheelchair system is shown in Figure~\ref{fig:previous work}. VAHM is equipped with a contact sensor, ultrasonic and infrared sensors to detect collisions, and wall-following algorithms to allow the wheelchair to navigate by following the wall. Wheelsley allows wheelchair users to give navigation commands by using an eye-tracking system. Vulcan1.0 uses 2D LiDAR and RGB camera input to build a visual 2D map for the wheelchair user to visualize their current location and their surroundings to let them better operate the wheelchair. The NavChair is an assistive wheelchair navigation system that integrates autonomous navigation, obstacle avoidance, door passage, and wall-following which allow the wheelchair to navigate autonomously. Vulcan2.0 is the improved version of Vulcan1.0 which integrates autonomous navigation algorithms and reduces the cost by using 2D LiDAR only. The Cyberwork Wheelchair is a commercially available smart wheelchair which is equipped with multiple RGB-D cameras to get the surrounding information and use it to navigate autonomously while avoiding obstacles. The performance comparison is shown in Table~\ref{tab: wheelchair comparison}. 

While these wheelchair platforms offer significant navigational assistance to individuals who struggle with traditional powered wheelchairs, they rely heavily either on sensor data or human input, which can have limitations and sensitivity to environmental changes~\cite{Leaman2016DevelopmentOA}. For user controlled wheelchair systems, such as those using eye-tracking control~\cite{Yanco1998Wheelesley}, they require substantial cognitive effort or fine motor skills, which can be fatiguing or unreliable over extended periods. Fully autonomous systems, on the other hand, can fall short in dynamic and unpredictable environments due to sensor limitations or algorithmic shortcomings~\cite{Sahoo2024}. Our proposed system is equipped with multiple sensors and a shared control-based navigation system that allow users to partially control the wheelchair which increases the reliability of the intelligent wheelchair system, addressing the limitations of fully autonomous systems in dynamic and unpredictable environments.

\section{Hardware Design}
\label{sec:hardware_design}
\par 

\begin{figure}[!tb]
    \centering
    \includegraphics[width =0.48\textwidth]{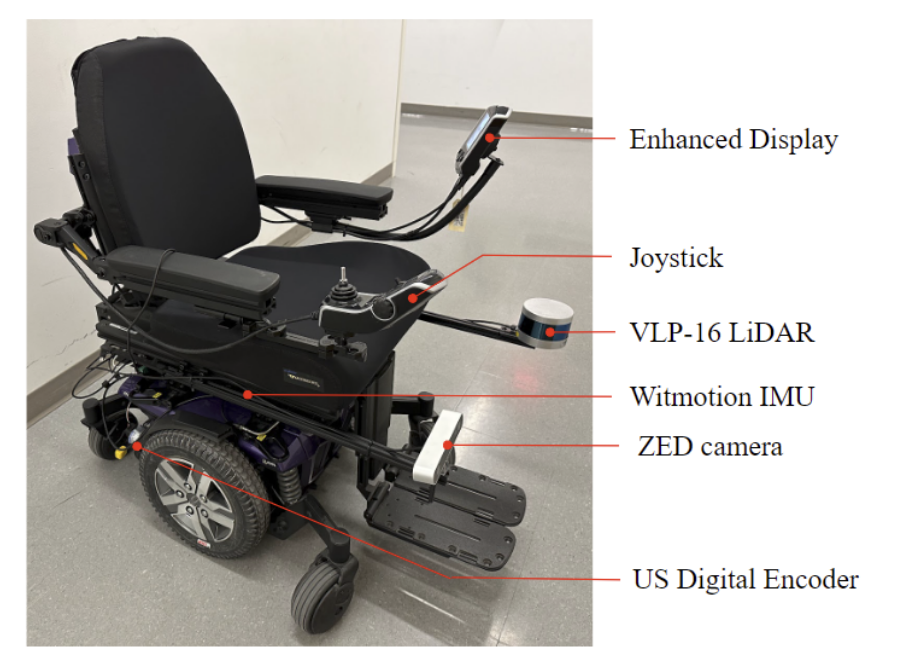}
    \caption{Hardware Design of the CoNav Chair}
    \label{fig:hardware}
\end{figure}

For this project, we have equipped a commercially available PWC, Quantum Q6 Edge 2.0, with control electronics, optical encoders, and multiple sensors such as VLP-16 LiDAR, ZED stereo camera, and Witmotion IMU. Sensors like LiDAR and cameras are basic for robot perception, mapping, and autonomous navigation. Unfortunately, most commercial PWC (e.g., Quantum Q6 Edge 2.0, Pride Mobility) are not equipped with the basic sensing system, and some smart wheelchairs in the market are still unaffordable for wide use. To develop a universal standard smart wheelchair, drawing from the inspiration of Turtlebot3~\cite{Amsters2020Turtlebot}, one of the most widely used differential drive mobile robots, we equipped our smart wheelchair with LiDAR, RGB-D camera, IMU, and encoders. 
Incorporating LiDAR provides precise distance measurement and environmental mapping, enabling accurate obstacle detection and avoidance. The camera adds visual perception, which is essential for recognizing and interpreting complex surroundings and dynamic obstacles. The inertial measurement unit (IMU) offers critical data on the wheelchair's orientation and movement, improving stability and control. The encoders contribute to precise wheel movement tracking, enhancing the overall accuracy of the wheelchair's navigation system. 
% Additionally, the closed-loop PID motor control module delivers precise and responsive motor control, which is crucial for smooth and reliable maneuverability. 
The hardware architecture is shown in Figure~\ref{fig:hardware}. 

\begin{figure}[!tb]
    \centering
    \includegraphics[width =0.5\textwidth]{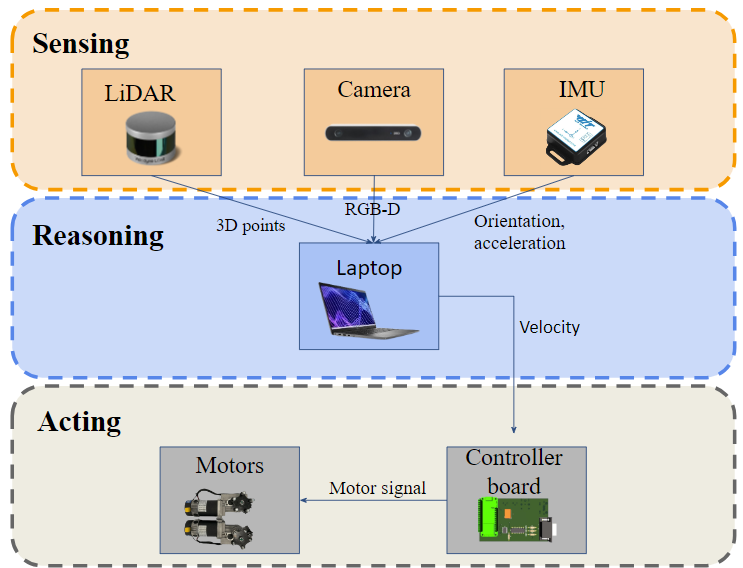}
    \caption{Hardware Dataflow of CoNav Chair}
    \label{fig:hardwaredata}
\end{figure}

As shown in Figure~\ref{fig:hardwaredata}, 
Our hardware structure follows the Sensing - Reasoning - Acting structure~\cite{Infantino2008}. For the Sensing part, all of the sensors, such as the LiDAR and the camera will provide the robot with environmental information about the surroundings and publish the sensor data to a laptop PC for further processing. For the Reasoning module, the laptop PC is connected to the sensors and motors, which run a Linux operating system and ROS. Our shared control-based navigation program will run on this PC and send actuator control commands to motors. For the Acting module, a motor control system based on R-Net is designed, and a closed PID control system is integrated to ensure the stability of the lower-level control system. 

\subsection{Motion Control Module}

\begin{figure}[!t]
    \centering
    \includegraphics[width =0.48\textwidth]{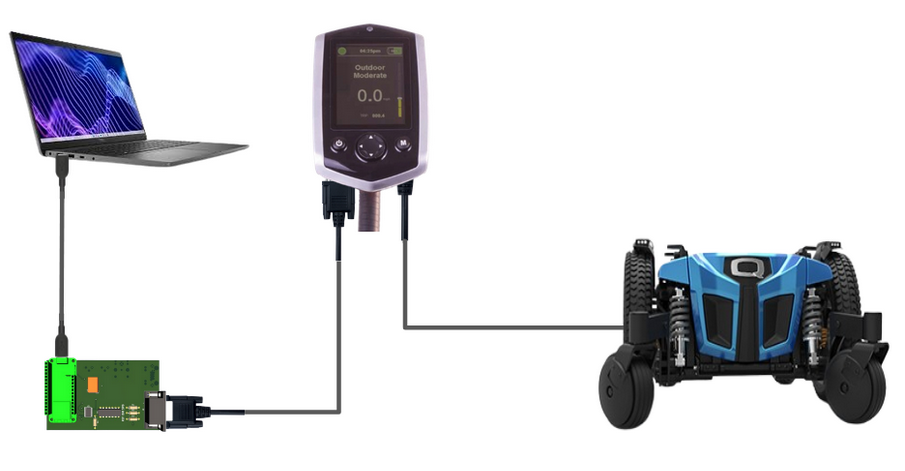}
    \caption{Motor Control Module}
    \label{fig:motor_control}
\end{figure}

For the motor control module, we are using the alternative control interface embedded in the wheelchair. The alternative control interface is on the enhanced display, which has a 9-way pin-out. We design our motor controller board, which can output two ways of voltage ranging from 4.8V - 7.2V to both control the forward and rotation velocity. The board is programmed to have a \textit{rosserial} module to take command velocity signals sent from our laptop by publishing and subscribing nodes and topic protocol. Inside the board, two digital Potentiometers are used to convert the control message to the voltage signal and send it to the alternative control interface. After the alternative control interface gets the voltage output from the controller board, the interface will convert the voltage signals to CAN messages and send them to the wheelchair base. The way we connect the motor controller module is shown in Figure~\ref{fig:motor_control}.  

\subsection{Sensor Module}

\begin{figure}[t]
    \centering
    \includegraphics[width =0.48\textwidth]{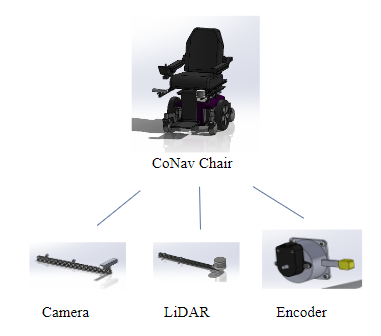}
    \caption{Sensor Module}
    \label{fig:Sensor Module}
\end{figure}

The LiDAR and camera were mounted on the wheelchair using T-slotted aluminum extrusions to avoid obstructing leg movement. The camera was secured via a pivot joint and M5 slide nuts, while the LiDAR was mounted using a bracket and M5 slide nuts. Both extrusions were attached to the wheelchair’s armrest and seat rail using L brackets and square nuts, as shown in Figure~\ref{fig:Sensor Module}. Cables were routed along the extrusions and secured with zip ties for a clean installation.

The wheelchair's motors were modified to retrofit new encoders while retaining the functionality of the integrated electromagnetic brakes. This involved preparing the motor shafts by drilling and threading them to accommodate an encoder mount. The encoders were securely installed using existing mounting points, following standard guidelines, to ensure precise and reliable operation.
\section{Software Design}
\label{sec:software_design}
\par 

\begin{figure*}[h!]
    \centering
    \includegraphics[width =0.85\textwidth]{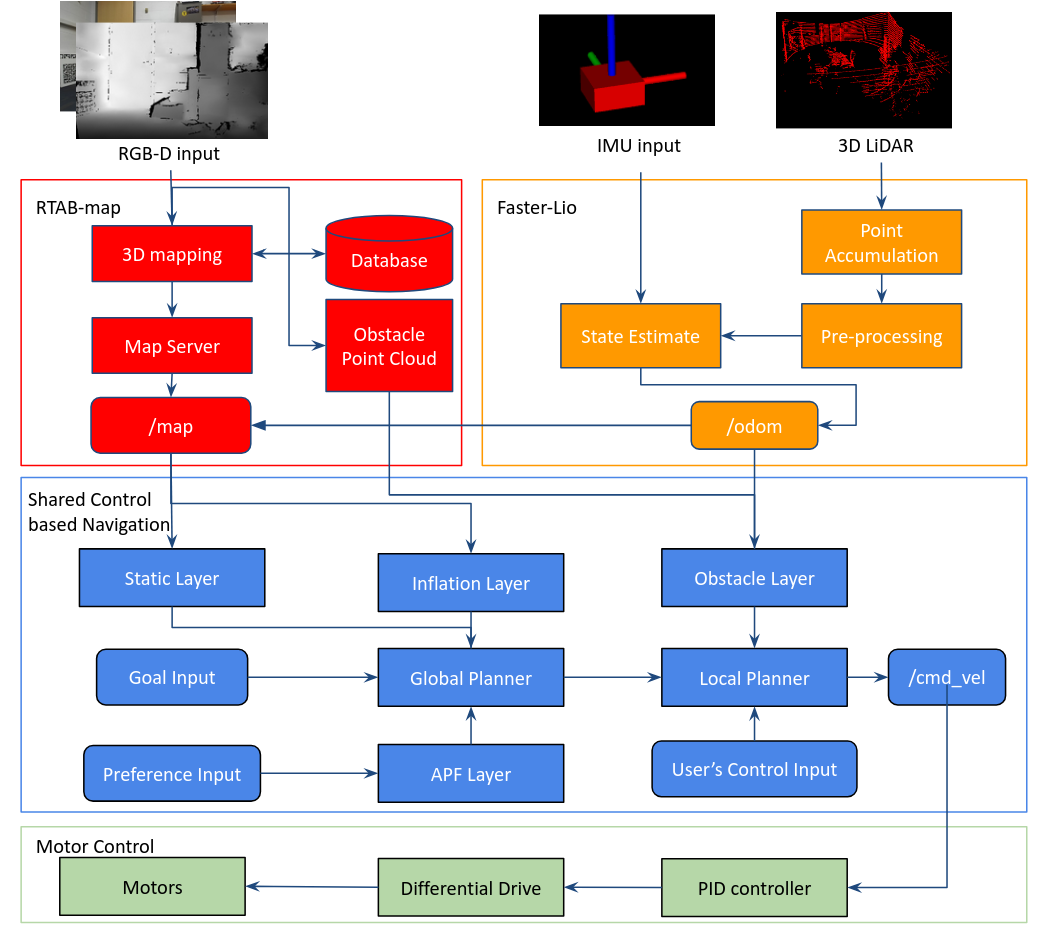}
    \caption{Software Structure of CoNav Chair}
    \label{fig:software}
\end{figure*}

Our proposed SLAM and shared control-based navigation system of the smart wheelchair is based on ROS~\cite{Quigley2009ROS}. ROS provides a robust and flexible framework for writing robot software, allowing developers to leverage a vast ecosystem of tools, libraries, and conventions designed for robot applications. The general structure of the software arrangement is shown in Figure~\ref{fig:software}. The structure consists of three parts: a 3D mapping module, a Localization module, a Shared control-based navigation module, and a Motor control module. 

\subsection{Mapper and Localizer}
For the mapping module, the system employs RTAB- Map~\cite{Labbe2013AppearanceBased}, which is a graph-based SLAM algorithm capable of generating both 3D maps and their 2D projections of the surrounding environment. The RGB-D sensor serves as the primary input to RTAB-Map, enabling the construction of detailed 3D maps through real-time point cloud data processing. This map data is stored in a database, which is also used for managing obstacle point clouds for effective collision avoidance. The RTAB-Map framework integrates with a map server to publish the map information, accessible under the \texttt{/map} topic, which is crucial for high-level path planning and navigation tasks.

Simultaneously, the system utilizes Faster-LIO~\cite{Bai2022FasterLIO}, a LiDAR-inertial odometry framework, for accurate localization. Faster-LIO processes data from the 3D LiDAR and IMU inputs. Initially, the raw LiDAR data undergoes pre-processing to filter out noise and structure the point cloud for efficient processing. These pre-processed point clouds are then accumulated to provide consistent spatial information. The state estimation module combines the IMU data with the processed LiDAR information, ensuring robust and precise localization even in dynamic environments. The resulting odometry data is published under the \texttt{/odom} topic, providing real-time position and orientation feedback to the wheelchair system.

\subsection{Shared Control-based Navigation}
\begin{figure}[h!]
    \centering
    \includegraphics[width = 0.5\textwidth]{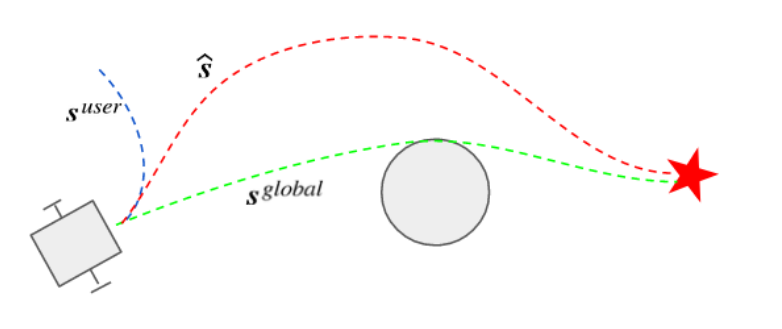}
    \caption{Shared Control-based MPC}
    \label{fig:share control}
\end{figure}
% By using the predefined map, the wheelchair user can set the desired location for the system to generate the path and navigate the wheelchair to its final destination. During the navigation process, we implement a shared control-based navigation structure, in which the wheelchair user can customize the global path in the user preference-based global planner and adaptively apply their control on the navigation process. This integration allows the system to combine the user's control input signals with the sensor data, enabling the wheelchair to navigate through the environment efficiently and safely. The shared control approach ensures that the user retains control over the wheelchair while benefiting from the system's autonomous navigation capabilities. This balance enhances user trust and promotes a seamless interaction between the user and the wheelchair.

Considering a wheelchair’s role as an intimately user-integrated robotic system, it becomes essential to incorporate considerations of the user’s control. Therefore, within our current framework, we have incorporated methods and concepts of shared control. In order to combine both user’s control and autonomous navigation, we proposed a Shared Control-based Model Predictive Control local planner~\cite{xu2024} to allow users to adaptively combine their control signals into the whole navigation process. 

Here we utilized a non-linear model predictive control (MPC)~\cite{Wang2021GroupBased} formulation for indoor environments navigation. The optimization problem is formulated below:

\begin{equation}
    \label{eq:vinalla}
    \begin{aligned}
    \mathbi{u}^{*} = \arg&\min_{\mathbi{u}\in\mathcal{U}}\mathcal{J}(\mathbi{s}, \hat{\mathbi{s}}, \mathbi{u})\\
    s.t. s_{t+1} &= f(s_{t}, u_{t}) \\
    \mathbi{s} &\in \mathcal{S}_{free} \\
    \mathbi{u} &\in \mathcal{U}
    \end{aligned}
\end{equation}

where $s = \{s_0,...,s_T\}$  is the state sequence of the robot drawn from the feasible set $S_{free}$by passing a control sequence $u=\{u_0,...,u_{T-1}\}$  drawn from a control space U through the robot dynamic $f$. $\hat{s}=\{\hat{s}_0,...,\hat{s}_T\}$ is the combined reference states of robot by considering both the user's control signal sequence and the global path plan generated by the user-preference field. $\mathcal{J}$ is the cost function that expresses the considerations of efficiency, safety, and user control. 

To enable the user to take control of the navigation process, we incorporate the predicted state generated by the user's control sequence into the cost function $\mathcal{J}$, as shown below.

\begin{equation}
    \label{eq: shared MPC}
    \begin{aligned}
        \mathcal{J}(\mathbi{s}, \hat{\mathbi{s}}, \mathbi{u}) &= \mathcal{J}_s(\mathbi{s}, \hat{\mathbi{s}}) + \mathcal{J}_{u}(\mathbi{u}) \\
        \mathcal{J}_s(\mathbi{s}, \hat{\mathbi{s}}) &= (\mathbi{s}-\hat{\mathbi{s}})^{T}\mathbi{Q}_{s}(\mathbi{s}-\hat{\mathbi{s}}) \\
        \hat{\mathbi{s}} = \eta(k)&\mathbi{s}^{user} + (1-\eta(k))\mathbi{s}^{global} \\
        \mathcal{J}^{u} &= \mathbi{u}^{T}\mathbi{Q}_{u}\mathbi{u}
    \end{aligned}
\end{equation}

where $\mathcal{J}_s$ represents the state cost and $\mathcal{J}_u$ represents the input cost. The combined cost, denoted as $\hat{s}$, takes into account both the global planning state sequence $\mathbi{s}_{global}$ and the user’s control sequence $\mathbi{s}_{user}$. The parameter $\theta(k) \in [0, 1]$ acts as a weight for the user’s control sequence, determining the degree of control the user exerts. The variable k indicates the number of control signals the user provides within a given time frame. To adjust the levels of user control in our system, we devised a weight function based on the frequency of user control, reflecting their intent to steer the wheelchair. An exponential function is used to model the user's intent, as shown below:

\begin{equation}
    \label{eq: shared MPC}
    \begin{aligned}
        \theta(k)=1-e^k
    \end{aligned}
\end{equation}

As shown in Figure~\ref{fig:share control}, the reference path combined the user’s control signal and the global path plan. The MPC local planner will then allow the robot to follow the combined path while avoiding obstacles.

\section{Experiment Setup and Result}
\label{sec:result}
\par 

In this part, we set up a real-world experiment and tested our wheelchair within an indoor-built environment. The experiment setup and the result are shown below.

\subsection{Experiment Setup}

\begin{figure}[t]
    \centering
     % \begin{subfigure}[b]{0.48\textwidth}
         \centering
         \includegraphics[width=0.48\textwidth]{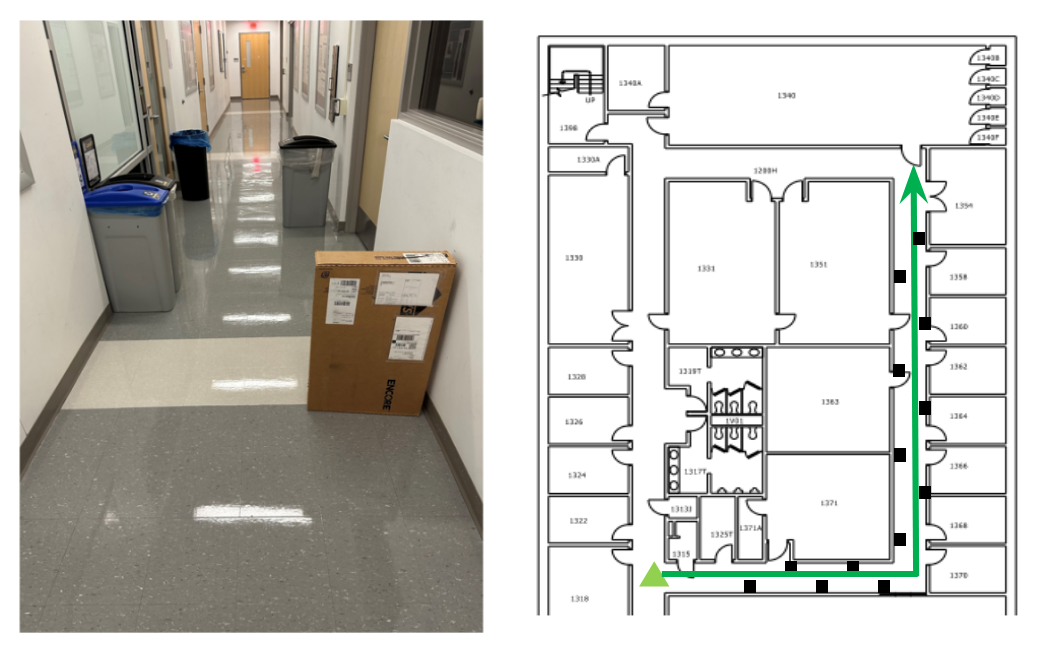}
         \caption{Experiment Environment and Navigation Route}
         \label{fig:environment}
     % \end{subfigure}
     % \begin{subfigure}[b]{0.48\textwidth}
     %     \centering
     %     \includegraphics[width=\textwidth]{figure/obstacle_avoidance.png}
     %     \caption{Illustration of SLAM and Obstacle Avoidance }
     %     \label{fig:obstacle}
     % \end{subfigure}
        % \caption{Experiment Setup}
        % \label{fig:Experiment Setup}
\end{figure}

\begin{figure*}[h!]
    \centering
    \includegraphics[width =0.8\textwidth]{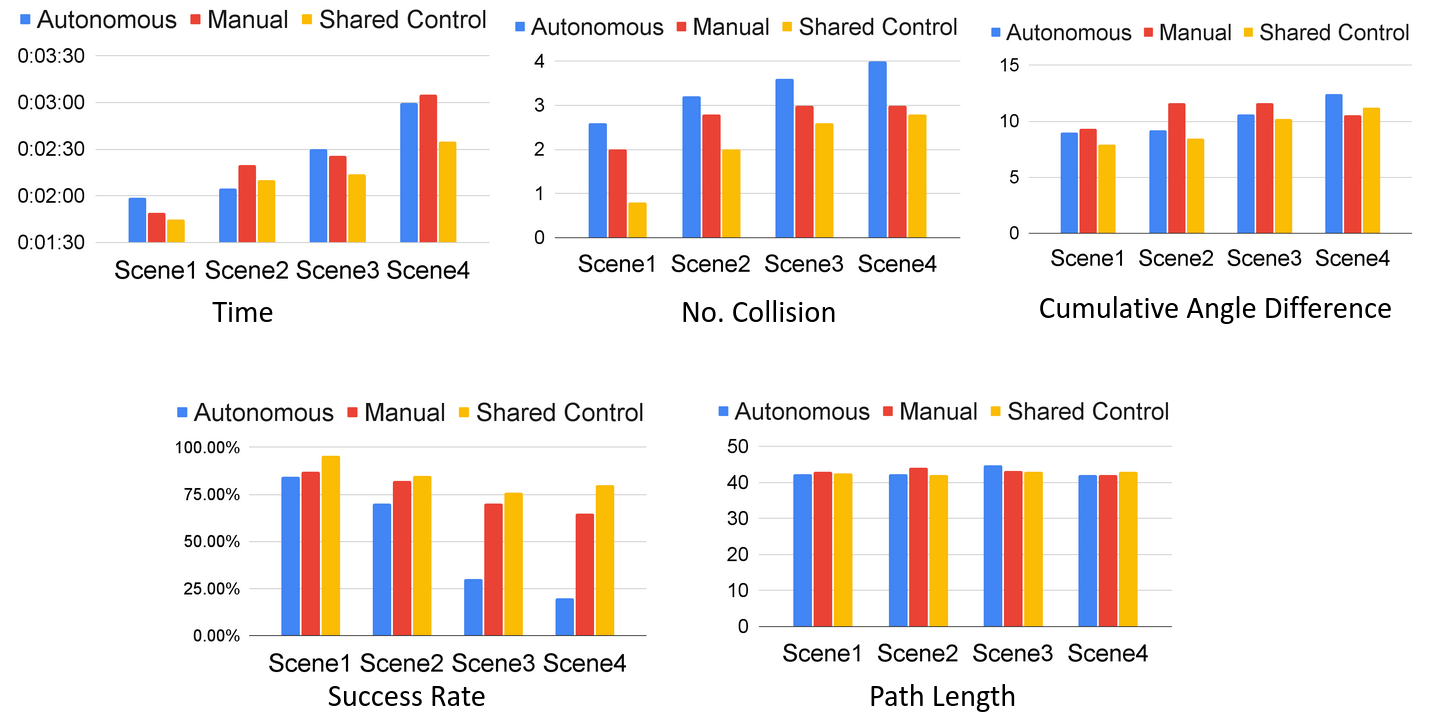}
    \caption{Quantitative Results of the CoNav Chair Evaluation Experiments}
    \label{fig:result}
\end{figure*}

The CoNav Chair runs on Ubuntu 20.04 with ROS Noetic on board. The navigation task is set up in the corridor on the first floor of an academic building on our campus, shown in Figure~\ref{fig:environment}. We create four corridor scenes where several obstacles are distributed around. One of the scene for testing is shown as Figure~\ref{fig:environment} This scene is used to test the performance of our shared-control algorithm compared with fully autonomous navigation and a fully manual control algorithm. We assessed the planner based on time, success rate, collision rate, user satisfaction, trajectory length, and angle difference summary representing smoothness~\cite{guillen2020} against fully manual and autonomous controls. 

 \subsection{Experiment Result}

The quantitative results in Figure~\ref{fig:result} show that the proposed shared-control-based framework significantly outperforms fully manual and autonomous navigation modes in both efficiency and reliability. The shared control planner achieves shorter completion times and trajectory lengths, reflecting faster and more direct navigation. Its smoother trajectories, indicated by lower cumulative absolute angle differences, ensure fewer abrupt turns and more natural directional changes—critical for applications like assistive robots where comfort and predictability are key. Additionally, the shared control planner demonstrates superior reliability, with higher success rates and fewer collisions across all scenarios.

Manual control, while allowing users to choose direct routes, often suffers from increased collisions and lower success rates due to users' unfamiliarity with the robot's dynamics and excessive adjustments near obstacles. Autonomous navigation, on the other hand, excels at obstacle avoidance and path optimization but struggles in complex scenarios like sharp turns or centrally positioned obstacles, often leading to inefficiencies or 'freezing robot' issues~\cite{trautman2010}. The shared control framework combines the strengths of both approaches. It leverages autonomous navigation in simpler environments while enabling user intervention in complex scenarios, allowing for effective handling of challenging conditions. This hybrid strategy provides a balanced solution, enhancing efficiency, smoothness, and overall reliability.
\section{Conclusion}
\label{sec:conclusion}
\par 

In this work, we propose the CoNav Chair, a ROS-based smart wheelchair with shared control-based navigation and obstacle avoidance that addresses critical gaps in current PWC designs, by overcoming the limitations of fully autonomous or manual systems. Combining a multi-sensor hardware platform with LiDAR, a camera, an IMU, and encoders for real-time situational awareness and a ROS-based shared control software framework, our system integrates user inputs with autonomous navigation for smoother and more efficient mobility. Tested in real-world built environments, the CoNav Chair demonstrates superior performance in navigation effectiveness and user comfort, offering a reliable and intuitive solution to enhance independence for PWD.
\section{Acknowledgement}
\label{sec:acknowledgement}

The work presented in this paper was supported financially by the United States National Science Foundation (NSF) SCC-IRG 2124857. The support of the NSF is gratefully acknowledged.

\bibliography{ISARC}

\end{document}